\documentclass[preprints,article, accept, pdftex,moreauthors]{Definitions/mdpi} 

\firstpage{1} 
\pubvolume{1}
\issuenum{1}
\articlenumber{0}
\pubyear{2025}
\copyrightyear{2025}
\datereceived{ } 
\daterevised{ } 
\dateaccepted{ } 
\datepublished{ } 
\hreflink{https://doi.org/} 

\Title{LLM-Powered Prediction of Hyperglycemia and Discovery of Behavioral Treatment Pathways from Wearables and Diet}

\TitleCitation{Title}

\Author{Abdullah Mamun $^{1,2,*}$\orcidA{}, Asiful Arefeen $^{2}$, Susan B. Racette $^{2}$, Dorothy D. Sears $^{2}$, Corrie M. Whisner $^{2}$, Matthew P. Buman $^{2}$, and Hassan Ghasemzadeh $^{2}$}

\AuthorNames{Abdullah Mamun, Asiful Arefeen, Susan B. Racette, Dorothy D. Sears, Corrie M. Whisner, Matthew P. Buman, and Hassan Ghasemzadeh}

\isAPAStyle{%
       \AuthorCitation{Mamun, A., Arefeen, A., \& Lastname, F.}
         }{%
        \isChicagoStyle{%
        \AuthorCitation{Lastname, Firstname, Firstname Lastname, and Firstname Lastname.}
        }{
        \AuthorCitation{Mamun, A.; Arefeen, A.; Racette, S.B.; Sears, D.D.; Whisner, C.M.; Buman, M.P.; Ghasemzadeh, H.}
        }
}

\address{%
$^{1}$ \quad School of Computing and Augmented Intelligence, Arizona State University, Tempe, AZ 85281, USA\\
$^{2}$ \quad College of Health Solutions, Arizona State University, Phoenix, AZ 85054, USA}

\corres{Correspondence: a.mamun@asu.edu}

\abstract{Postprandial hyperglycemia, marked by the blood glucose level exceeding the normal range after consuming a meal, is a critical indicator of progression toward type 2 diabetes in people with prediabetes and in healthy individuals. A key metric for understanding blood glucose dynamics after eating is the postprandial area under the curve (AUC). Predicting postprandial AUC in advance based on a person's lifestyle factors, such as diet and physical activity level, and explaining the factors that affect postprandial blood glucose could allow an individual to adjust their lifestyle accordingly to maintain normal glucose levels. In this study, we developed an explainable machine learning solution, GlucoLens, that takes sensor-driven inputs and uses advanced data processing, large language models, and trainable machine learning models to predict postprandial AUC and hyperglycemia from diet, physical activity, and recent glucose patterns. We used data obtained from wearables in a five-week clinical trial of 10 adults who worked full-time to develop and evaluate the proposed computational model that integrates wearable sensing, multimodal data, and machine learning.  Our machine learning model takes multimodal data from wearable activity and glucose monitoring sensors, along with food and work logs, and provides an interpretable prediction of the postprandial glucose pattern. Our GlucoLens system achieves a normalized root mean squared error (NRMSE) of 0.123 in its best configuration. On average, the proposed technology provides a 16\% better performance level compared to the comparison models. Additionally, our technique predicts hyperglycemia with an accuracy of 73.3\% and an F1 score of 0.716 and recommends different treatment options to help avoid hyperglycemia through diverse counterfactual explanations. Code available: https://github.com/ab9mamun/GlucoLens.}


\usepackage{amsmath,amssymb,amsfonts}
\usepackage{algorithmic}
\usepackage{graphicx}
\usepackage{textcomp}
\usepackage{xcolor}
\usepackage{multirow}

\usepackage{caption}
\usepackage{subcaption}
\usepackage{float}
\usepackage{subfloat}
\usepackage{comment}

\newcommand{\figref}[1]{Fig.~\ref{fig:#1}}
\newcommand{\secref}[1]{Section~\ref{sec:#1}}
\newcommand{\tblref}[1]{Table~\ref{tbl:#1}}

\newcommand{\eqnref}[1]{Equation~\ref{eqn:#1}}

\begin{document}

\maketitle


\section{Introduction}
\label{sec:intro}
Wearable sensors have become indispensable tools in the field of healthcare and personalized medicine, offering continuous, non-invasive monitoring of physiological parameters in real time \cite{vaghasiya2023wearable, clubbs2023use}. Devices such as continuous glucose monitors (CGMs), physical activity trackers, and smartphone sensors provide valuable insights into an individual’s metabolic health, physical activity, and lifestyle habits \cite{uddin2021human, mamun2022multimodal}. These sensors enable the collection of granular data that can be used to understand complex physiological processes, predict health outcomes, and design interventions \cite{shields2024continuous, steck2019continuous}. Wearable sensors also provide opportunities for developing robust solutions despite different challenges, such as small dataset, missing features, data in the wild, etc. \cite{mamun2022designing, maray2023transfer, mamun2025aimi, Soumma_Mamun_Ghasemzadeh_2025}. In the context of diabetes management, wearable sensors allow for the precise tracking of blood glucose fluctuations and physical activity levels, offering individuals and clinicians the means to implement timely and tailored strategies for disease prevention and management \cite{pickup2015real}. This accessibility to continuous health data marks a significant shift toward proactive and data-driven healthcare solutions.

Hyperglycemia, or high glucose concentration in the blood, occurs when the body cannot effectively regulate glucose levels. Hyperglycemia in the fasted state (i.e., no caloric intake for at least 8 hours) is defined by a fasting plasma glucose level $\geq$ 100 mg/dL ($\geq$5.6 mmol/L), whereas hyperglycemia in the non-fasted state generally is defined as having a blood glucose level (BGL) $\geq$140 mg/dL (7.8 mmol/L) two hours after a meal \cite{nihTypeDiabetes,gehlaut2015hypoglycemia}. Lack of physical activity and relative overconsumption of carbohydrates can affect a person's metabolism and their ability to regulate glucose \cite{crichton2015physical}. Other common risk factors often cited as potentially responsible for developing diabetes are being obese, overweight, or having a higher than normal body mass index (BMI) \cite{maggio2003obesity}. Untreated hyperglycemia increases the risk of complications like retinopathy, nephropathy, neuropathy, cardiovascular disease, stroke, poor limb circulation, and depression \cite{mouri2023hyperglycemia}. While hyperglycemia can affect anyone, individuals with prediabetes or diabetes are at a higher risk compared to healthy people \cite{alvarez2017prediabetes,castro2022mayo}.

\begin{figure}[t!]
    \centering
    \includegraphics[width=0.70\textwidth]{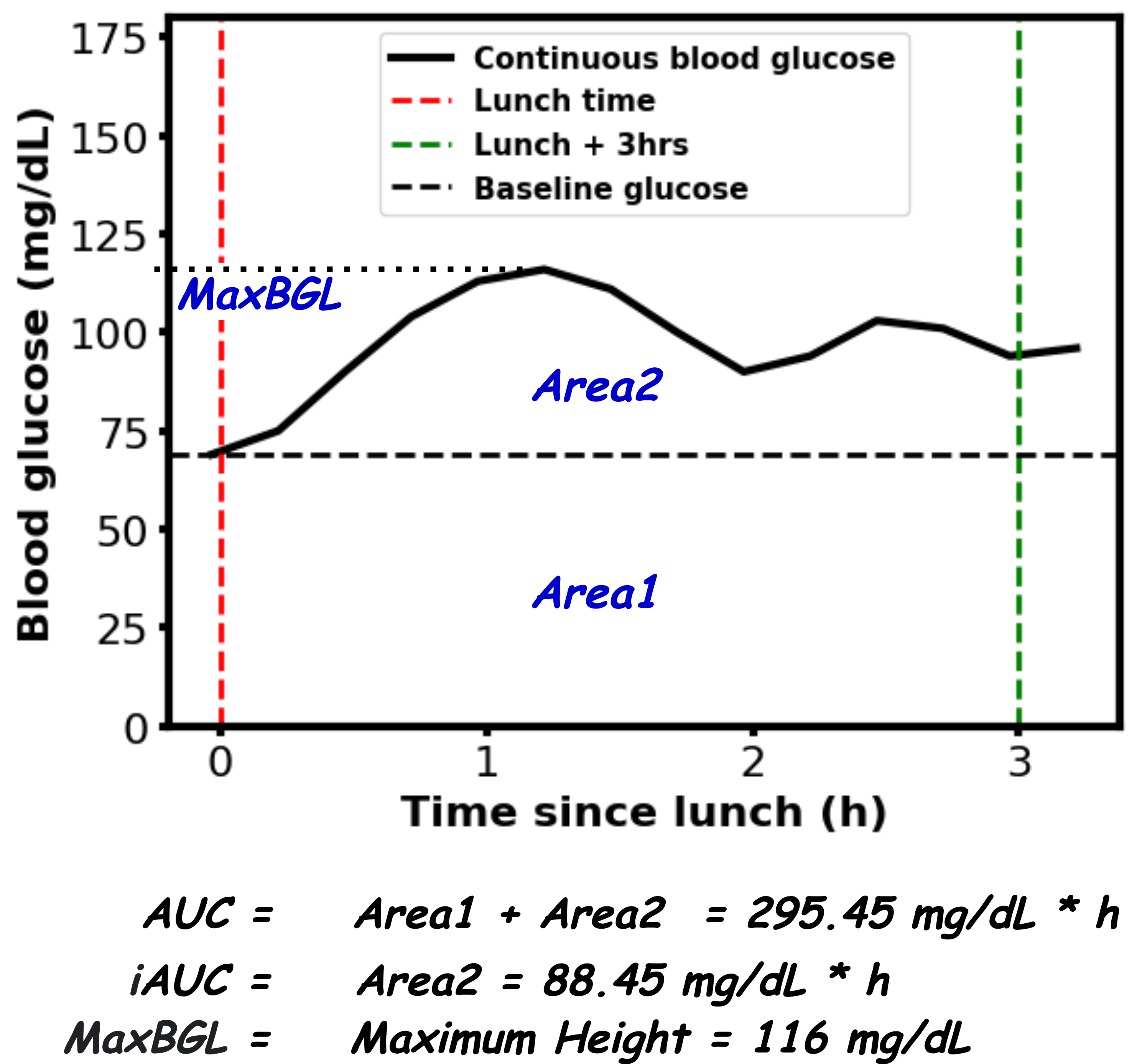}
    \caption{An example of postprandial Area Under the Curve (AUC), incremental AUC (iAUC), and Maximum postprandial Blood Glucose Level (MaxBGL) for one participant after a lunch meal.}
    \label{fig:auc_example}
\end{figure}

The Centers for Disease Control (CDC) estimates that 38\% of American adults are prediabetic and 19\% of them are unaware of their condition \cite{cdcNationalDiabetes}. To help prevent the increasing prevalence of hyperglycemia and prediabetes, the Food and Drug Administration (FDA) has approved the sale of Continuous Glucose Monitors (CGMs) over the counter in the U.S. in 2024 \cite{fdaClearsFirst}. This decision has made CGM devices more accessible to people with or without diabetes. Prediabetes can be reversed with proper lifestyle management, such as diet and physical activity \cite{echouffo2023diagnosis}. However, if left untreated, it can develop type 2 diabetes, which is an irreversible lifelong condition \cite{chakarova2019assessment}. Healthy individuals are generally expected to maintain their blood glucose levels (BGL) within a range of 60 to 140 mg/dL \cite{nihTypeDiabetes}. However, temporary elevations above this range are common following certain meals. The area under the BGL curve during a specific post-meal period (e.g., 2–3 hours) is referred to as the postprandial Area Under the Curve (AUC), which is an important indicator of healthy BGL regulation. Numerous studies have explored methods for calculating AUC and the various factors influencing it \cite{floch1990blood, sakaguchi2016glucose, t2023breaking}. An example of calculating AUC, incremental AUC, and the maximum postprandial blood glucose level (MaxBGL) is shown in \figref{auc_example}.

Blood glucose forecasting using wearable sensor data is an active area of research that has been addressed with different machine learning and deep learning approaches, including ensemble methods, attention methods, and knowledge distillation \cite{han2024glu, farahmand2024hybrid, farahmand2025attengluco}. Despite the increasing interest in the area under the glucose curve as a metric to estimate the risk of hyperglycemia \cite{uemura2017comparison, ugi2016evaluation}, to the best of our knowledge, the use of diet, physical activity, work routine, and previous baseline glucose parameters to predict hyperglycemia and the area under the glucose curve remains unexplored.

In this paper, we leverage wearable sensor data and propose and develop a machine learning solution, GlucoLens, for estimating the postprandial AUC and predicting hyperglycemia with wearable sensor data from CGM and activity trackers, as well as food and work logs. To summarize, our contributions are as follows: (i) formulating a problem of predicting AUC and hyperglycemia in advance based on diet, work habit, and physical activity, (ii) designing hyperglycemia prediction systems based on machine learning and large language models that combines wearable data, food logs, work logs, and recent CGM patterns to make prediction, (iii) implementing GlucoLens for AUC and hyperglycemia prediction and diverse counterfactual (CF) explanations, (iv) train and test our models on a dataset collected in a clinical trial conducted by our team, (v) provide a thorough discussion of our findings and how it advances the research of hyperglycemia prediction.

\section{Materials and Methods}

\begin{figure}[t!]%
    \centering
    \subfloat[Simplified workflow of the GlucoLens system]{\includegraphics[width=0.75\linewidth]{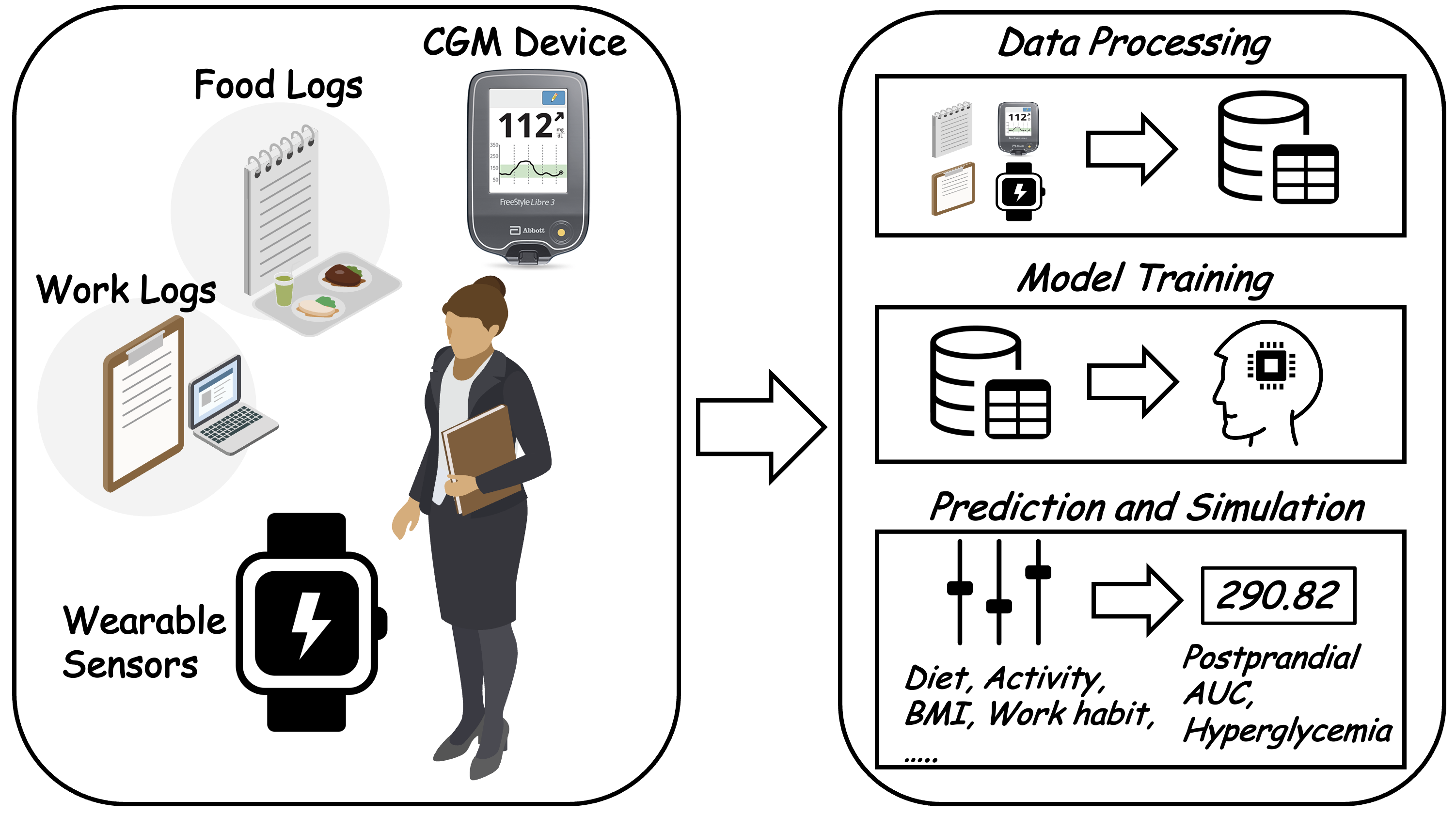}}%
    \\
    \subfloat[Detailed pipeline of the GlucoLens system]{\includegraphics[width=0.99\linewidth]{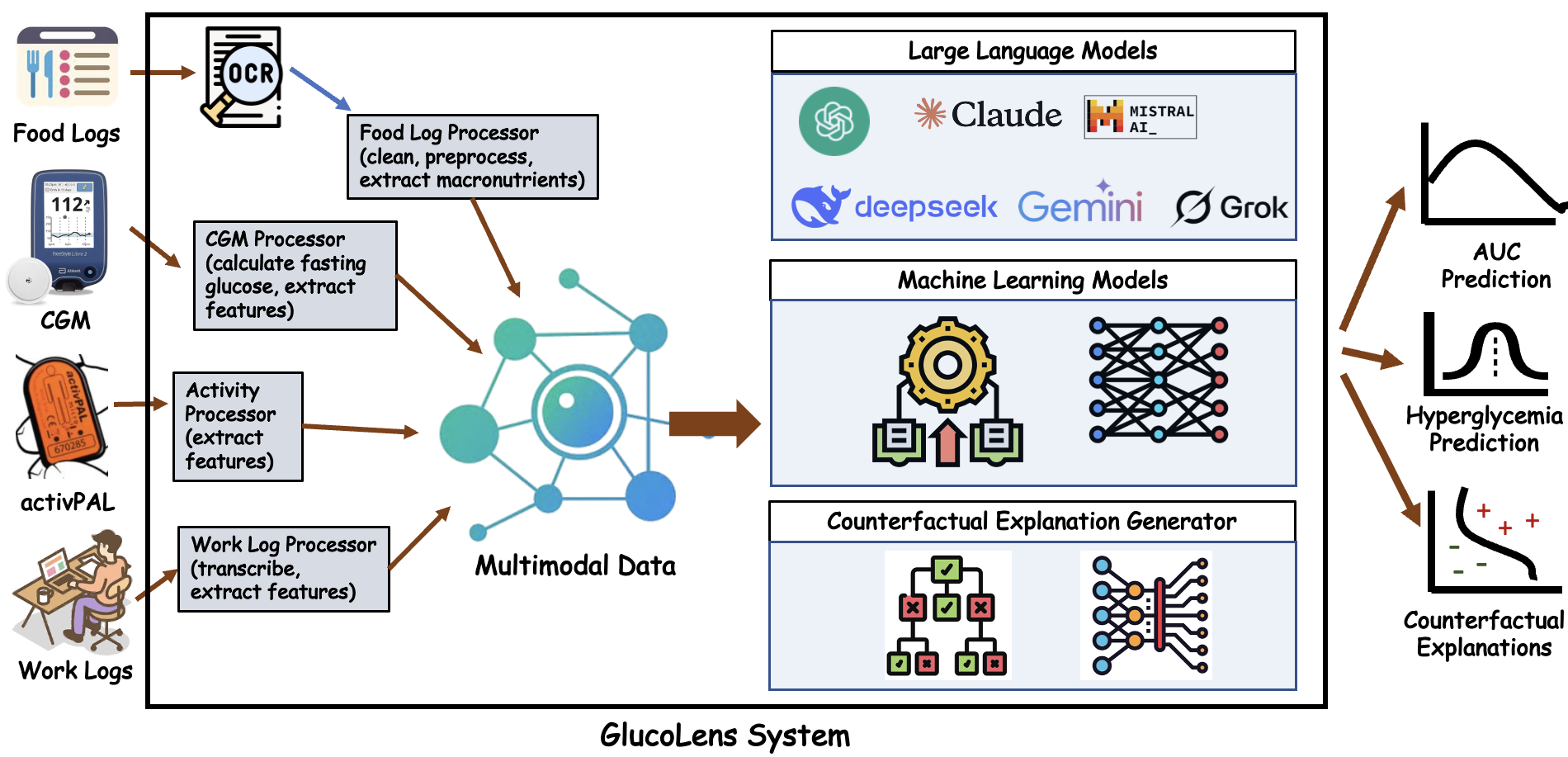}}%
    \caption{The GlucoLens system's body sensor network is composed of food logs, work logs, a CGM device, and a wearable sensor. Data from all these sources are combined to create a unified dataset to train a model. The machine learning models are further supported by large language models (LLM) and data augmentation tools based on Gaussian noise. Finally, counterfactual explanations are provided so that interventions can be applied.}
\label{fig:system_diagram}
\end{figure}

\subsection{System Overview}
To investigate the questions about the relationships of diet and activity with postprandial BGL, we designed a system of four data sources: a wearable device with motion sensors, a CGM device, food logs, and work logs. The system's data processing tools process the raw data from multiple sources and convert them to compatible electronic formats. The processed data are then used to train ML models which are capable of making accurate predictions about the AUC and hyperglycemia given the type of diet and activity. The system is illustrated in \figref{system_diagram}.

\begin{figure}[h]%
    \centering
    \hspace{0.5mm}
    \subfloat[activPAL micro device, to be worn on thigh]{\includegraphics[width=0.25\linewidth]{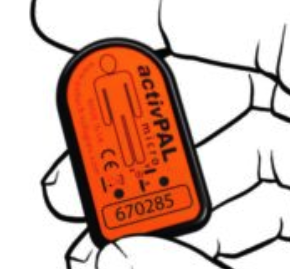}}%
    \hspace{2mm}
    \subfloat[Freestyle Libre CGM patch and reader]{\includegraphics[width=0.25\linewidth]{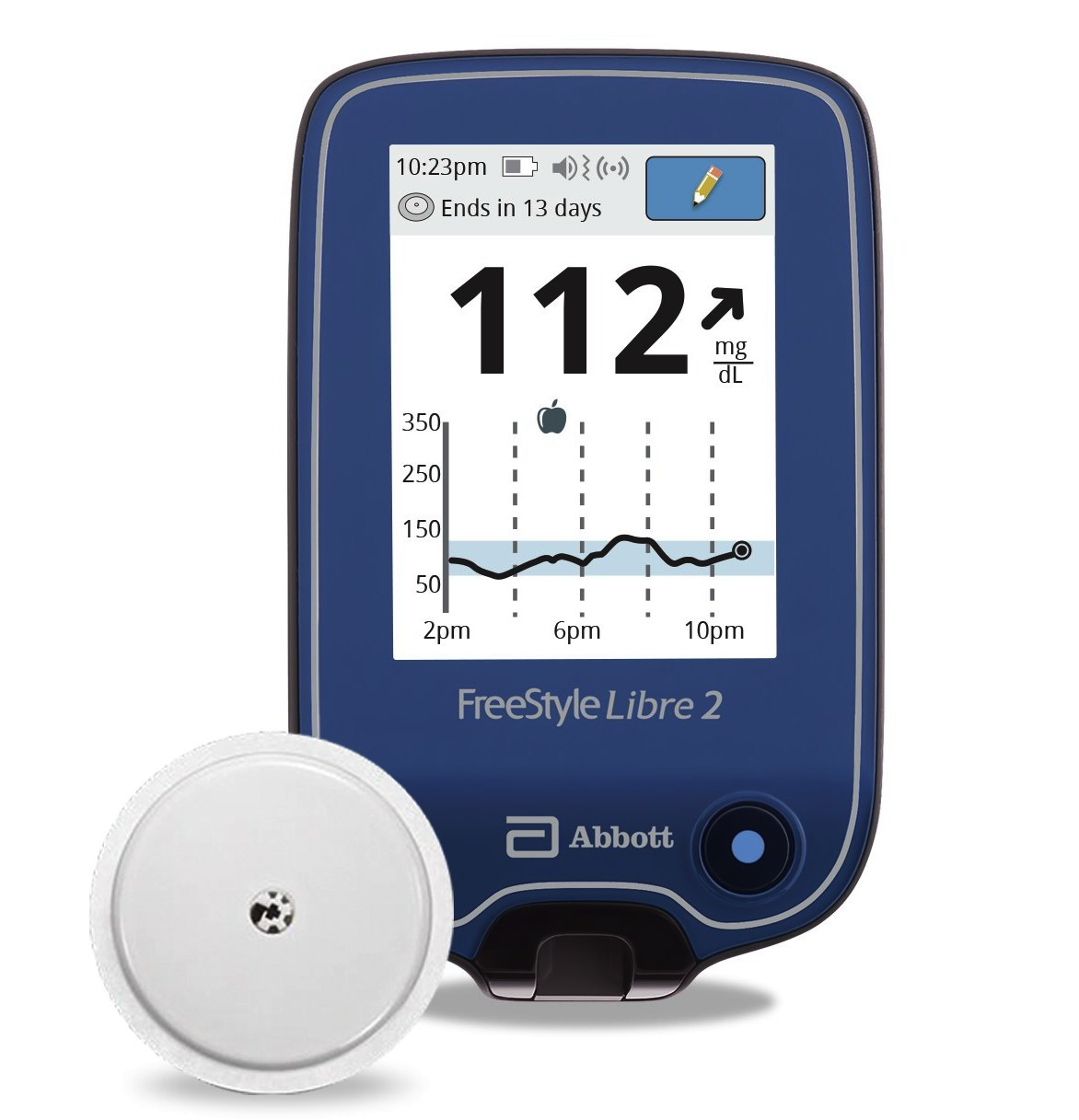}}%
    \hspace{2mm}
    \subfloat[GENEActiv wearable wristband]{\includegraphics[width=0.22\linewidth]{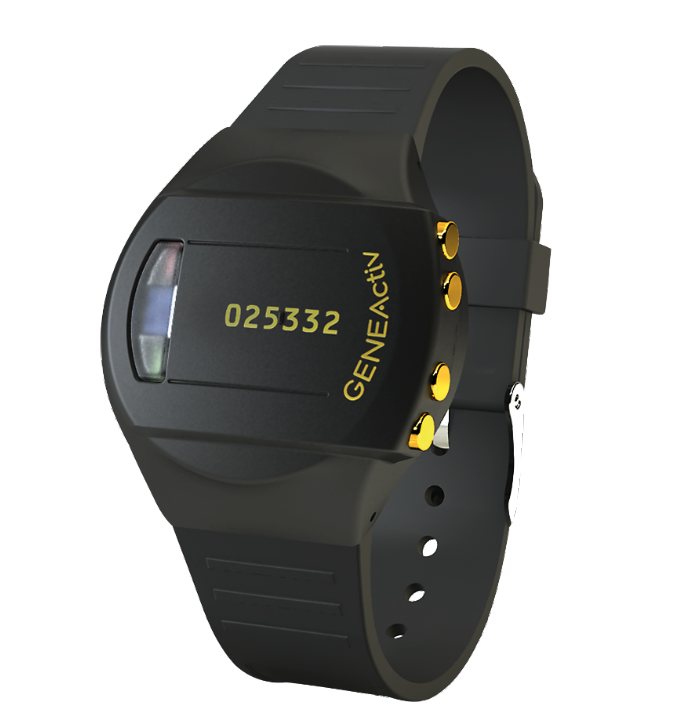}}%
    \caption{Wearables with sensors in the WorkWell Study. activPAL uses a tri-axial accelerometer for detecting activities. The FreeStyle Libre CGM patch measures glucose levels in the interstitial fluid (the fluid surrounding cells) using a thin filament inserted under the skin. The GENEActiv device uses a 3-axis accelerometer, a light sensor (photodiode), and a temperature sensor.}
\label{fig:sensors}
\end{figure}

The system has three wearable devices: a FreeStyle Libre CGM device, an activPAL micro device, and a GENEActiv wearable wristband, as shown in \figref{sensors}. activPAL provides event data with the timestamps of that specific event \cite{paltEventsExtended}. Changes in activity are considered as events by the activPAL device, e.g, when a person stands up from a sedentary position or starts moving from a standing position are considered events. The different activities or events detected by the device are: sedentary, standing, stepping, cycling, primary lying, secondary lying, and seated transport. For each event, the event start time and the duration of the event are recorded. Duration spent in different activities can be calculated from those events. The Freestyle Libre provides CGM readings in 15 minute-intervals. GENEActiv provides various biomarkers including acceleration, physical activity intensity, sedentary vs movement activity, sleep wake time \cite{activinsightsGENEActivData}.

GlucoLens is composed of a multimodal data processing package, interfaces for an array of large language models (LLMs), trained machine learning models curated for postprandial AUC and hyperglycemia predictions, and a counterfactual explanation generator. The data processing package includes Optical Character Recognition (OCR), food log processor, CGM processor, activity processor, and work log processor. Our system processes the raw data from the sensors and these sources and prepare a multimodal dataset. The multimodal data is used to produce predictions from 7 different LLMs: GPT-3.5-Turbo, GPT-4, Claude Opus 4, Deepseek V3, Gemini Flash 2.0, Grok 3, and Mistral Large. The Mistral Large model declined to provide predictions, hence omitted from the results, and was not considered for the hybrid predictors. Additionally, machine learning backbones such as Random Forest (RF), Ridge Regressor, Multilayer Perceptron (MLP) Regressor, as well as XGBoost \cite{chen2016xgboost}, and TabNet \cite{arik2021tabnet}. The multimodal dataset is augmented by adding small Gaussian noise to the scaled features. The system then produces predictions for postprandial AUC, maximum blood glucose level, and postprandial hyperglycemia. Finally, diverse counterfactual explanations are provided so that interventions can be designed to avoid hyperglycemia.

\subsection{Wearables and Lifestyle Data}
\label{sec:user_study}
    

\begin{figure}
    \centering
    \includegraphics[width=0.90\linewidth]{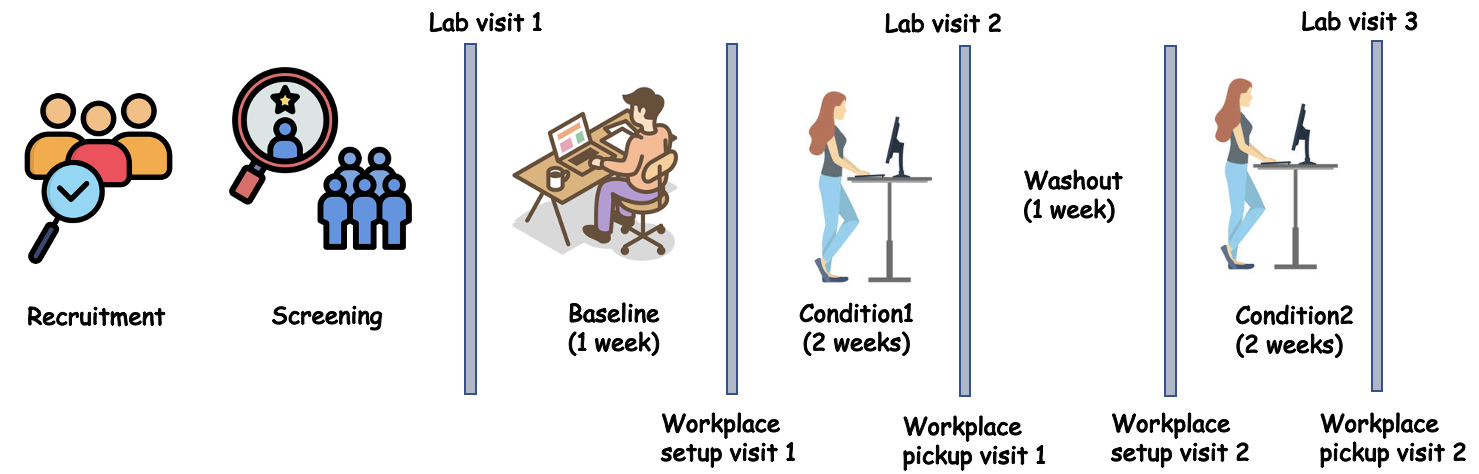}
    \caption{Timeline of the WorkWell Study. Each participant completed a 1-week Baseline phase data collection, 2 weeks of Condition1 phase data collection, and 2 weeks of Condition2 phase data collection. There was a 1-week break between Condition 1 and Condition 2. In all three phases, participants provided CGM, activPAL, GENEActiv, food logs, and work logs data. The Baseline phase was the control phase, where the participants maintained their usual work habits. Participants were randomly assigned an intervention of `Stand' or `Move' in Condition 1, and the other intervention was assigned in Condition 2. In the `Stand' intervention, they were told to stand as much as possible at work, and during the `Move' intervention, they were told to move as much as possible at work.}
    \label{fig:study_timeline}
\end{figure}

To develop the proposed system, we used a subset of data obtained in the WorkWell Study \cite{wilson2023effects}, a clinical trial (NCT04269070) conducted by our team that involved adults who worked full time. The WorkWell Study was approved by the Arizona State University Institutional Review Board, and all participants provided written informed consent to participate. Each participant was given a CGM device and an activPAL wearable \cite{lyden2017activpal} device. Lunch was delivered at work from select restaurants on every working day (five days per week) for one work week at Baseline, two work weeks during Condition 1, and two work weeks during Condition 2, as shown in \figref{study_timeline}. Each participant received printed forms to maintain the food logs and work logs. The option of homemade meals was not considered for lunch on working days because it would make the task of estimating the macronutrients difficult. The selected restaurants already had the macronutrients for all their items available online, which was used to estimate the nutritional components for every meal consumed.

In the food logs, the participants noted down the time and what they ate in every meal, including if there were any leftovers. The participants also logged the amount of water and any supplements (e.g., multivitamins) they were taking every day. In the work logs, they noted when they started working, when they stopped working, how they arrived at work or if they were working from home, and approximately what percentage of working time they spent sitting, standing, and walking.

The Baseline week did not have any intervention but the participants wore their CGM devices and activPAL. Two interventions for Condition 1 and Condition 2 were `recommending to be standing' and `recommending to be moving' as much as possible during work. All participants were provided with both interventions, but the temporal order was random. Therefore, for some participants, the temporal order was the Baseline, Stand, Move, and for the other participants, it was the Baseline, Move, Stand. In this study, we have included our analysis based on the data of 10 participants having an average Baseline BMI of $32.8 \pm 4.5$.

\subsection{Data Processing}

\begin{table}[t!]
\caption{A comparison of the five different feature sets used in this study. The checkboxes indicate whether a feature is present in a specific feature set. GL = glycemic load of the meal. Self = activity scores calculated by self-reported activity duration in work logs. Macro = macronutrients that are used to calculate glycemic load: net carb, fat, protein, and carb. Sensor = activity metrics from the activPAL sensor: duration of sitting, standing, and stepping activities of the day before lunch and the same metrics from the start of working to just before lunch of the same day.}
\centering
\begin{tabular}{lllllll}
\hline
\textit{No.} & \textit{Feature name/} & \textit{Sensor} & \textit{Sensor} & \textit{Self} & \textit{Self} & \textit{All} \\
& \textit{shorthand} & \textit{+GL} &  \textit{+Macro}& \textit{+GL}&\textit{+Macro} & \\
\hline 
1 & Fasting glucose & $\boxtimes$ & $\boxtimes$ & $\boxtimes$ & $\boxtimes$ & $\boxtimes$ \\
2 & Recent CGM & $\boxtimes$ & $\boxtimes$ & $\boxtimes$ & $\boxtimes$ & $\boxtimes$ \\
3 & Lunch time & $\boxtimes$ & $\boxtimes$ & $\boxtimes$ & $\boxtimes$ & $\boxtimes$  \\
4 & Work from home & $\boxtimes$ & $\boxtimes$ & $\boxtimes$ & $\boxtimes$ & $\boxtimes$  \\
5 & BMI & $\boxtimes$ & $\boxtimes$ & $\boxtimes$ & $\boxtimes$ & $\boxtimes$    \\
6 & Calories & $\boxtimes$ & $\boxtimes$ & $\boxtimes$ & $\boxtimes$ & $\boxtimes$  \\
7 & Calories from fat & $\boxtimes$ & $\boxtimes$ & $\boxtimes$ & $\boxtimes$ & $\boxtimes$    \\
8 & Saturated fat & $\boxtimes$ & $\boxtimes$ & $\boxtimes$ & $\boxtimes$ & $\boxtimes$   \\
9 & Trans fat & $\boxtimes$ & $\boxtimes$ & $\boxtimes$ & $\boxtimes$ & $\boxtimes$   \\
10 & Cholesterol & $\boxtimes$ & $\boxtimes$ & $\boxtimes$ & $\boxtimes$ & $\boxtimes$    \\
11 & Sodium & $\boxtimes$ & $\boxtimes$ & $\boxtimes$ & $\boxtimes$ & $\boxtimes$   \\
12 & Total carbs & $\boxtimes$ & $\boxtimes$ & $\boxtimes$ & $\boxtimes$ & $\boxtimes$    \\
13 & Sugar & $\boxtimes$ & $\boxtimes$ & $\boxtimes$ & $\boxtimes$ & $\boxtimes$   \\
14 & Work start time & $\boxtimes$ & $\boxtimes$ & $\boxtimes$ & $\boxtimes$ & $\boxtimes$    \\
15 & Day of the week & $\boxtimes$ & $\boxtimes$ & $\boxtimes$ & $\boxtimes$ & $\boxtimes$    \\ 
16 & activPAL &$\boxtimes$ & $\boxtimes$ & $\square$ & $\square$ & $\boxtimes$  \\
17 & Self reported acitivity & $\square$ & $\square$ & $\boxtimes$  & $\boxtimes$ &  $\boxtimes$ \\
18 & GL & $\boxtimes$ & $\square$ & $\boxtimes$ & $\square$ &  $\boxtimes$  \\
19 & Net carbs & $\square$ &$\boxtimes$  & $\square$ & $\boxtimes$ &   $\boxtimes$ \\
20 & Fat & $\square$ & $\boxtimes$ & $\square$ & $\boxtimes$ &  $\boxtimes$  \\
21 & Protein & $\square$ & $\boxtimes$ & $\square$ & $\boxtimes$ & $\boxtimes$    \\
22 & Fiber & $\square$ & $\boxtimes$ & $\square$ &$\boxtimes$  & $\boxtimes$   \\ \hline
\end{tabular}
\label{tbl:feature_sets}
\end{table}

In this study, one of the main interests was predicting postprandial blood glucose during work days and investigating how diet and activity during work days affect postprandial AUC. Moreover, the participants were provided standardized lunch on working days, whereas they ate anything they liked on weekends. In the food logs, the participants kept track of any leftover portions, which allowed us to accurately extract the amount of calories and macronutrients consumed during lunch. The food logs were maintained as handwritten logs. They were processed through Google Cloud Vision OCR to create electronic logs, followed by some human intervention for issues that could not be resolved by the OCR. The amounts of macronutrients consumed were used to estimate glycemic loads (\textit{GLs}) using the formula from \cite{lee2021development} as shown in \eqnref{glycemic_load}.

\begin{equation}
\begin{aligned}
GL = 19.27 &+ (0.39 \times \text{net carb.}) - (0.21 \times \text{fat})\\ & - (0.01 \times \text{protein}^2) - (0.01 \times \text{fiber}^2)
\end{aligned}
\label{eqn:glycemic_load}
\end{equation}

The work logs were also handwritten, and they were processed manually to convert them to an electronic format. From activPAL sensors data and with the help of work logs, the durations of sitting, standing, and stepping were calculated for the day to lunch, as well as the durations of sitting, standing, and stepping during work hours for the day until lunch. For this work, fasting glucose was defined as the minimum CGM reading between 6 AM and 10 AM. The recent CGM was defined as the average CGM reading of the same day from midnight to 8 AM. A complete list of the input features of the different feature sets that were formed can be found in \tblref{feature_sets}. The `Sensor+Macro' feature set uses activity metrics from the sensor data and calculated macronutrients, along with the common features that are present in all feature sets. Two additional inputs are used in the `All' feature set: an activity score calculated from self-reported activity logs during work and the glycemic load. In the `Self+Macro' feature set, the 6 features containing sitting, standing, and stepping durations are replaced by the activity score based on self-reported activity logs. The activity score was calculated from the self-reported activities in the work logs. In the work logs, each participant reported the percentage of their working hours spent sitting, standing, and walking. The recent activity score is calculated by taking the average percentage of time spent in walking activity in the previous days of the same phase and adding with $\frac{1}{2} \times$ the average percentage of time spent in standing activity in the previous days of the same phase. In the `Self+GL' and `Sensor+GL' feature sets, the macronutrients: net carbs, fat, protein, and fiber are removed and replaced with the glycemic load calculated in \eqnref{glycemic_load}. The relationship of all five different feature sets can be found in \tblref{feature_sets}.

\subsection{ML Models}

\begin{table}[t!]
\centering
\caption{An overview of the different backbones and their hyperparameters for the regression problem. Explanations of the feature sets can be found in \tblref{feature_sets}. MLP = multilayer perceptron. RF = random forest. Ridge = ridge regression. $n_{est}$ = number of estimators in random forest. Meanings of the outcomes AUC and MaxBGL can be found in \figref{auc_example}. Gly\_Hybrid = best performing regressor (RF) with support of LLM-based predictions, Gly\_Hybrid\_v2 is RF with LLM-based predictions but with only the best LLM (Claude Opus 4). Gly\_Max is Gly\_Hybrid\_v2 with augmented training data with Gaussian noise.}
\begin{tabular}{ll}
\hline
\textit{Target outcomes}      & AUC,  MaxBGL, Hyperglycemia                             \\ \hline
\textit{Feature }        & Sensor + Macro, Self + Macro, \\ 
sets & Sensor + GL, Self + GL, All \\ \hline
               & RF, Ridge, MLP, XGBoost, TabNet, GPT-3.5, GPT-4, Mistral Large,\\
\textit{Predictors} & Gemini Flash 2.0, Claude Opus 4, Grok 3, Deepseek V3, Gly\_Hybrid,\\
 & Gly\_Hybrid\_v2, Gly\_Max, Hybrid Predictors for Classification \\
 & (RF+MLP, RF+XGB, XGB+MLP, RF+XGB+MLP).                                         \\ \hline
\textit{Ridge variations}     & $\alpha \in \{1, 0.1, 0.01\}$                             \\ \hline
\textit{RF variations}        & $n_{est} \in \{10, 50, 100\}$                                \\ \hline
\textit{MLP variations}       & 13 variations; see \tblref{mlp_hyperparams}                                            \\ \hline
           
\end{tabular}
\label{tbl:all_experiments}
\end{table}

\begin{table}[b!]
    \centering
    \caption{Variations of the multilayer perceptron (MLP regressor). Different variations have been tested by varying the depth and size of each layer.}
    \begin{tabular}{lll}
    \hline
\textit{Variation} & \textit{\# Hidden} & \textit{\# Nodes in}                           \\
\textit{no.}       & \textit{layers}    & \textit{hidden layers}                     \\ \hline
1         & 3         & (20, 10, 5)                       \\
2         & 4         & (40, 20, 10, 5)                   \\
3         & 4         & (60, 30, 15, 7)                   \\
4         & 5         & (80, 40, 20, 10, 5)               \\
5         & 5         & (100, 50, 25, 12, 6)              \\
6         & 5         & (120, 60, 30, 15, 7)              \\
7         & 5         & (140, 70, 35, 17, 8)              \\
8         & 5         & (160, 80, 40, 20, 10)             \\
9         & 8         & (80, 40, 20, 20, 20, 20, 10, 5)   \\
10        & 8         & (100, 50, 25, 25, 25, 25, 12, 6)  \\
11        & 8         & (120, 60, 30, 30, 30, 30, 15, 7)  \\
12        & 8         & (140, 70, 35, 35, 35, 35, 17, 8)  \\
13        & 8         & (160, 80, 40, 40, 40, 40, 20, 10) \\ \hline
    \label{tbl:mlp_hyperparams}
    \end{tabular} 
\end{table}

\begin{figure}
    \centering
    \includegraphics[width=0.99\linewidth]{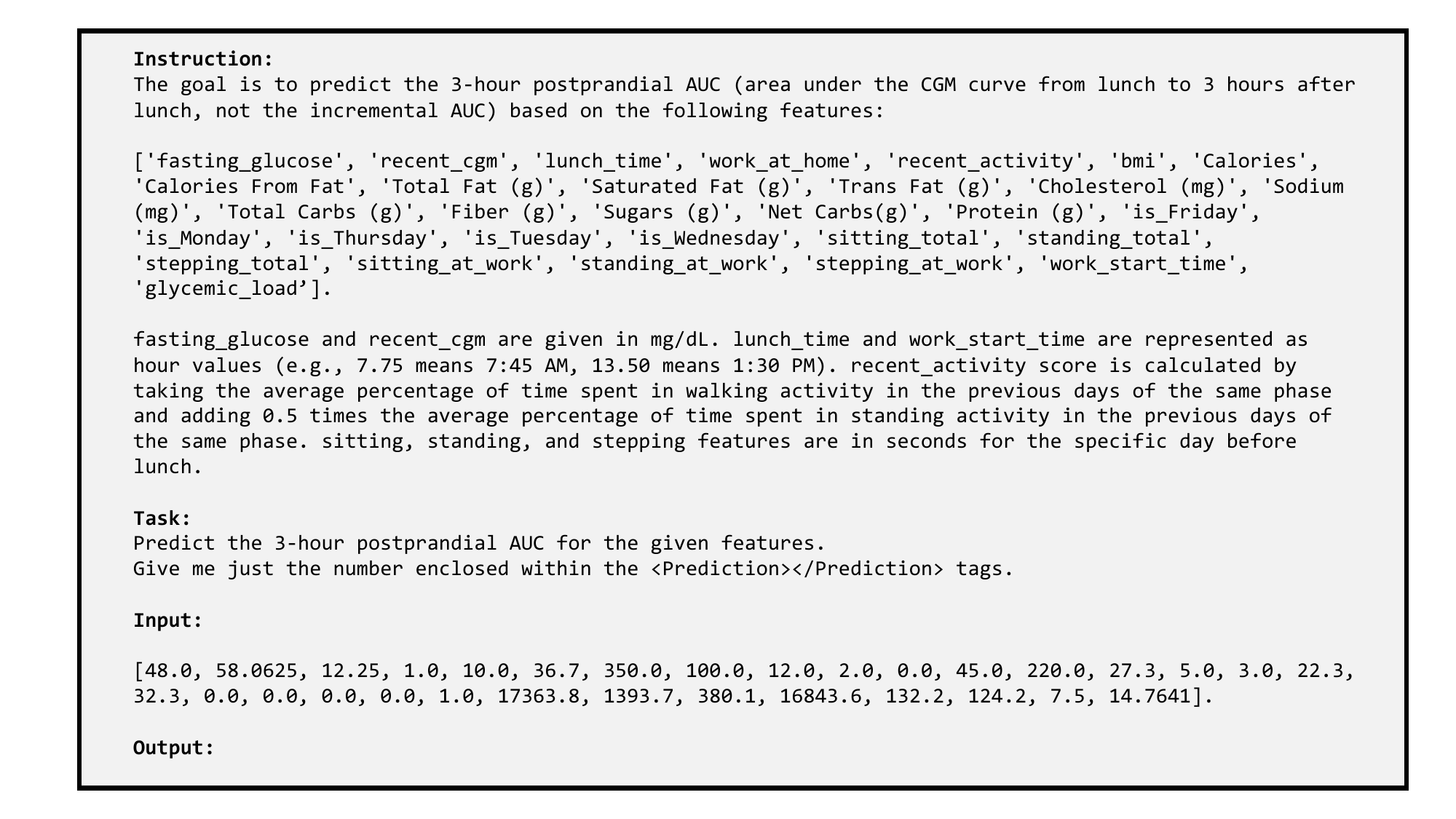}
    \caption{A sample zero-shot prompt for the LLMs. The input field is varied for each different example; everything else in the prompt remains the same.}
    \label{fig:llm_prompt}
\end{figure}

Five different architectures: Random Forest (RF), Ridge Regressor, MLP Regressor, XGBoost, and TabNet, were trained as potential backbone architectures of the GlucoLens system. MLPs and Random Forests are capable of making effective predictions with proper feature extractions \cite{azghan2025cudle, mamun2023neonatal}, whereas linear regression methods with regularization, such as Ridge Regression \cite{mcdonald2009ridge} and Lasso Regression \cite{ranstam2018lasso}, are popular choices as comparison models. Neural networks are usually data-hungry, which is why, on small datasets, classical ML methods can show competitive and even better performance than neural networks.

Additionally, zero-shot large language models were used for LLM-only predictions based on the feature values. The structure of the prompt used for the LLMs is presented in \figref{llm_prompt}. The content for the "Input:" field is changed for each different datapoint, whereas the rest of the body of the prompt remains the same. In the next step, hybrid predictors that use both zero-shot LLM predictions and a trainable regressor (RF or XGBoost) were implemented. Three variations of the hybrid predictors were implemented and trained. The first variation (Gly\_Hybrid) uses the existing features and adds the predictions by six advanced LLMs as additional inputs. The second hybrid variation (Gly\_Hybrid\_v2) uses the existing features and the prediction by only the best-performing LLM (in our case, Claude Opus 4) for training and testing. Finally, the third hybrid variation (Gly\_Max) uses a Gaussian noise-based augmented feature set based on existing features, predictions by Claude Opus 4, and a trainable regressor.

Three variations of RF, three variations of Ridge, and thirteen variations of MLP, one version of XGBoost, and one version of TabNet were trained and tested for the \textit{Gly\_Base} version of the GlucoLens system. The different models and variations for AUC, MaxBGL, and Hyperglycemia prediction are presented in  \tblref{all_experiments} and \tblref{mlp_hyperparams}. For AUC and MaxBGL predictions, 20\% data instances were assigned to the test set, and 80\% data instances were assigned to the training set. For hyperglycemia detection, six different variations of train-test split were used, which are discussed in details in \secref{hyperglycemia_results} and \tblref{different_training_size}.

\subsection{Hyperglycemia Detection and Counterfactual Explanations}

\begin{figure}[t!]
    \centering
    \includegraphics[width=0.98\linewidth]{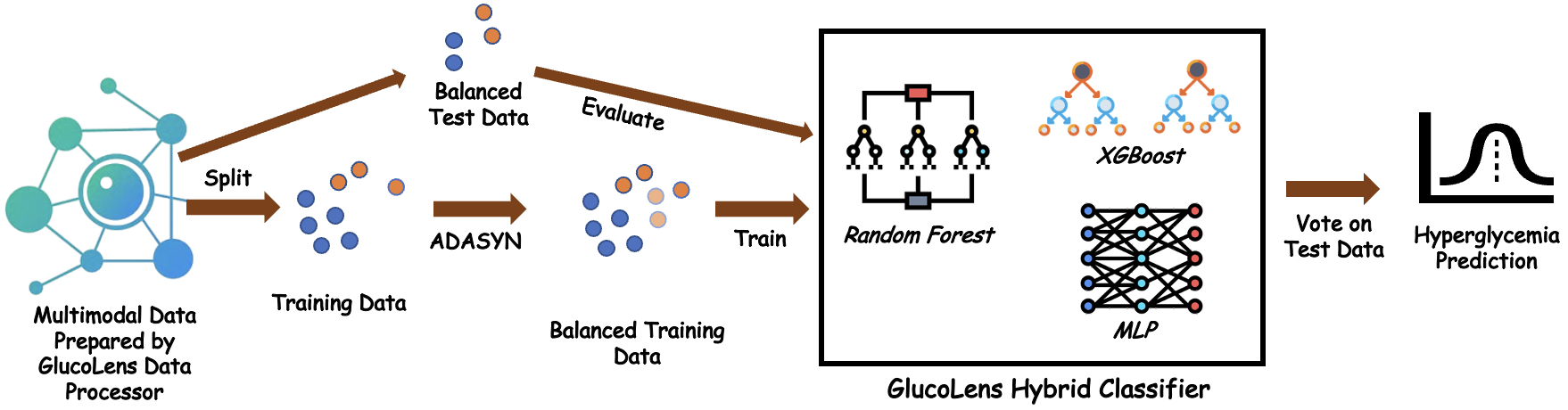}
    \caption{The pipeline of classification of hyperglycemia with GlucoLens system's hybrid classifier. A soft voting based on the probabilities of classes suggested by the RF, XGBoost, and MLP (version 13 of \tblref{mlp_hyperparams}) is used to make the final predictions. This hybrid method outperforms the prediction performances of single classifiers. The test set contains only real datapoints, so that the evaluation is not biased by the data generation method, whereas the balanced training data contains both real and synthetic datapoints.}
    \label{fig:classification_pipeline}
\end{figure}

GlucoLens was also trained for hyperglycemia detection with RF, XGB, MLP, backbone models, as well as hybrid combinations of these three, as shown in \figref{classification_pipeline}. More details about all the different variations of the models and different splits of training and test sets are described in \secref{hyperglycemia_results}. Counterfactual explanations can provide insights on features responsible for an undesired health outcome and possible remedies to overcome them \cite{mamun2024use, arefeenglyman}. As a primary goal of the paper was to explore more knowledge about the reasons for and ways to prevent hyperglycemia, a DiCE-based counterfactual explanations generator is integrated with GlucoLens. GlucoLens provides multiple counterfactual explanations that are diverse and achievable with small perturbations from the original example \cite{mothilal2020explaining}. 

\section{Results}
\label{sec:results}

\subsection{Prediction of AUC with Different Backbone Architectures}

The first problem we want to solve is the prediction of postprandial AUC and we used five different architectures as the backbone of our model, that includes 3 different versions of Random Forest (RF), 3 different versions of Ridge regressors, 13 versions of MLP, and finally, XGBoost \cite{chen2016xgboost} and TabNet \cite{arik2021tabnet}. According to our experiments, GlucoLens performs the best with an RF backbone with a normalized root mean squared error (NRMSE) of 0.123. TabNet trained with 100 epochs provided an NRMSE of 0.147, and XGBoost provided an NRMSE of 0.137. A summary of the performance of all these different backbones can be seen in \tblref{model_comparison}. We believe that advanced neural networks integrated with our system will be able to produce more accurate results, but it will require more data to train them. We show in \secref{hyperglycemia_results}, where the results of hyperglycemia detection are discussed, that the performance on the test set monotonically improves as we increase the size of the training dataset.

\begin{table}[htbp]
    \centering
    \caption{Normalized Root Mean Squared Errors (NRMSE) of our GlucoLens models (RF, Ridge, MLP, XGBoost, TabNet) for different feature sets in the prediction of postprandial AUC. Explanations of the feature sets can be found in \tblref{feature_sets}.}
    \begin{tabular}{llllll} 
    \hline
    \textit{Feature Set} & \textit{RF}    & \textit{Ridge} & \textit{MLP} & \textit{XGBoost} & \textit{TabNet} \\ \hline
    Sensor + GL          & \textbf{0.125} & 0.139          & 0.169        & 0.137            & 0.160           \\
    Sensor + Macro       & \textbf{0.123} & 0.142          & 0.172        & 0.139            & 0.147           \\
    Self + GL            & 0.142          & \textbf{0.139} & 0.178        & 0.152            & 0.154           \\
    Self + Macro         & \textbf{0.139} & 0.142          & 0.172        & 0.149            & 0.151           \\
    All                  & \textbf{0.123} & 0.140          & 0.176        & 0.137            & 0.151           \\ \hline
    \end{tabular}
    \label{tbl:model_comparison}
\end{table}

\subsection{Prediction with LLMs}

\begin{figure}
    \centering
    \includegraphics[width=0.90\linewidth]{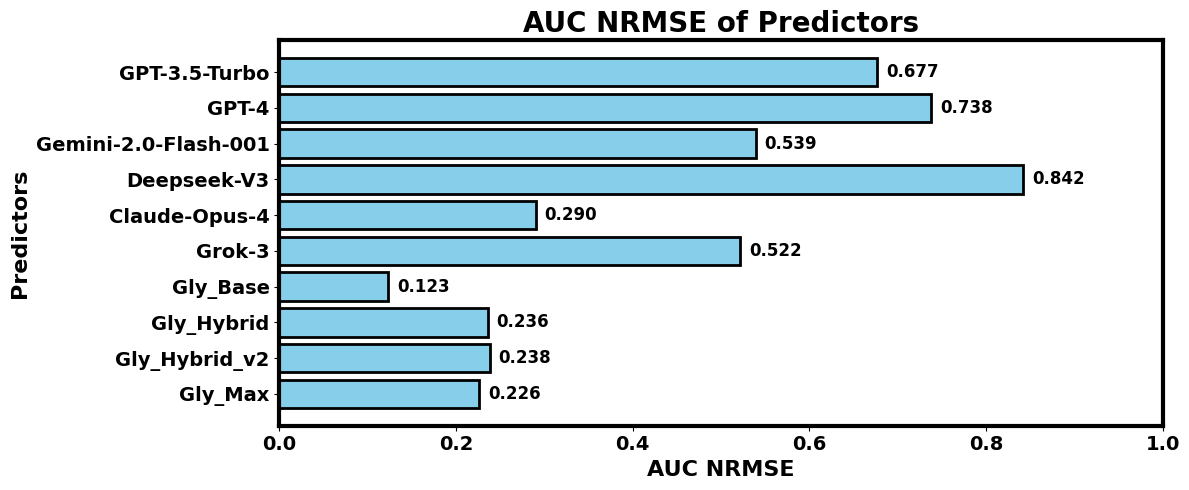}
    \caption{Normalized root mean squared errors on the prediction of AUC with different LLMs and hybrid classifiers of our system. Explanations of Gly\_Base and the hybrid predictors can be found in \tblref{all_experiments}. Explanations of the feature sets can be found in \tblref{feature_sets}.}
    \label{fig:llm_predictions}
\end{figure}

The results presented in Table~\ref{tbl:llm_results} and \figref{llm_predictions} showcase the performance of various configurations of our proposed solution. These configurations include \textit{Gly\_Base}, which does not utilize large language models (LLMs); \textit{Gly\_LLM}, a zero-shot prediction approach using only LLMs; \textit{Gly\_Hybrid}, combining LLM-based predictions with our system; \textit{Gly\_Hybrid\_v2}, which integrates predictions exclusively from the best-performing LLM (Claude); and \textit{Gly\_Max}, an enhanced version of \textit{Gly\_Hybrid\_v2} with augmented training data incorporating Gaussian noise. Performance of the hybrid models was evaluated by integrating LLMs with one of the two best-performing regressor backbones: Random Forest (RF) or XGBoost. Among the configurations, \textit{Gly\_Base} with RF backbone achieved the best performance. Interesting findings can be observed in \figref{llm_predictions}, where we can see that Claude Opus 4 outperformed all other LLM-based models by a large margin. These findings indicate that it may not be wise to ask LLMs for personalized meal plans as they cannot accurately predict the consequences of a certain combination of diet and other factors on the postprandial blood glucose levels. We used the `All' feature set for the LLM-based experiments.

\begin{table}[]
\centering
\caption{AUC NRMSE results of different variations of our solution. Gly\_Base = GlucoLens regressor with no LLM, Gly\_LLM = LLM only prediction (zero-shot) after multimodal data processing by GlucoLens. The hybrid predictors are explained in \tblref{all_experiments}.}
\begin{tabular}{llclll}
\toprule
\textit{Backbone} & \textit{Gly\_Base} & \multicolumn{1}{l}{\textit{Gly\_LLM}} & \textit{Gly\_Hybrid} & \textit{Gly\_Hybrid\_v2} & \textit{Gly\_Max} \\
\midrule
RF       & \textbf{0.123}     & \multirow{2}{*}{0.290}       & 0.241       & 0.238           & 0.226    \\
XGBoost  & 0.137     &                              & 0.236       & 0.242           & 0.259   \\
\bottomrule
\end{tabular}
\label{tbl:llm_results}
\end{table}

\subsection{Performance of MLP and Prediction of MaxBGL}
\begin{figure}[t!]
    \centering
    \includegraphics[width=0.90\linewidth]{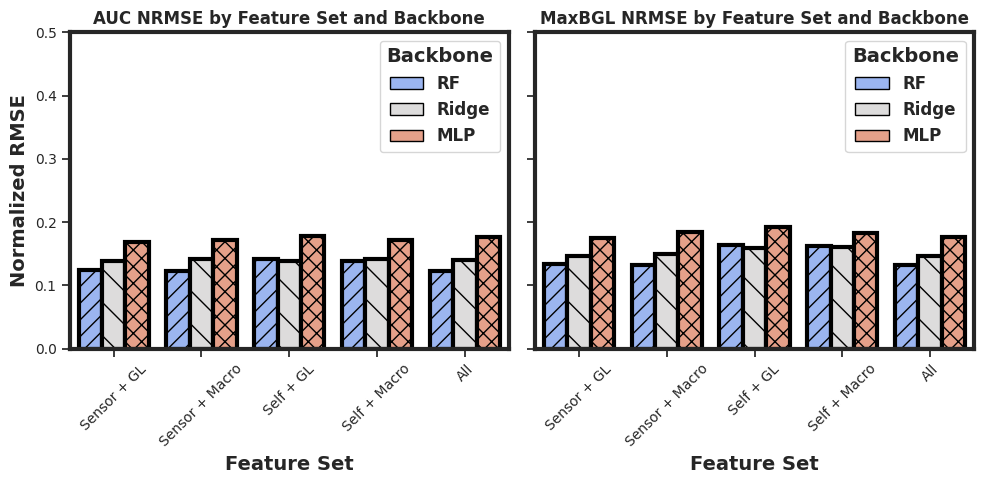}
    \caption{Summary of the best results for each target outcome with RF, Ridge, and MLP regressors. Three different models were tested with five different combinations. Meanings of the outcomes AUC and MaxBGL can be found in \figref{auc_example}. RF = random forest. NRMSE = normalized root mean squared error.}
    \label{fig:results_for_rf_ridge_mlp}
\end{figure}

Two different outcomes were chosen for the regression problem of predicting AUC and MaxBGL. In \figref{results_for_rf_ridge_mlp}, we see that for AUC and MaxBGL, the best NRMSE values are 0.123 and and 0.132, respectively. RF predictors obtained all three best results. Although 13 different variations were tested for MLPs, none of them achieved the best performance on any of the three tasks.

\begin{figure}[t!]
    \centering
    \includegraphics[width=0.9\linewidth]{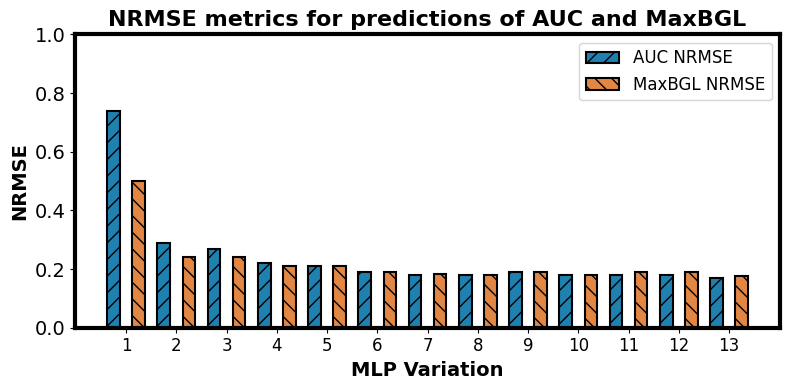}
    \caption{Normalized root mean squared errors (NRMSE) for different variations of MLP in two different regression tasks. Best NRMSE for AUC prediction is 0.169, and best NRMSE for MaxBGL prediction is 0.175. The details of the MLP variations can be found in \tblref{mlp_hyperparams}. The best performance among MLPs was achieved with the MLP variation no. 13 and the feature set `Sensor + GL'.}
    \label{fig:mlp_results_plot}
\end{figure}

Out of the 13 different variations of MLP regressors, all of their NRMSE metrics were higher than the corresponding RF regressors, as shown in \figref{results_for_rf_ridge_mlp}. It encouraged us to look deeper into the results of the MLP variations. We present the best performances by each of the 13 MLP variants for both outcomes in \figref{mlp_results_plot}. MLP variation no. 13 is the largest of all the MLP variations used in our experiments, and it performed better than any other MLP versions in both tasks. It raises the question of whether even larger and deeper models would be more accurate and potentially better than the RF and Ridge models. We wish to investigate this in future work.

\begin{figure}[h]%
    \centering
    \includegraphics[width=0.9\linewidth]{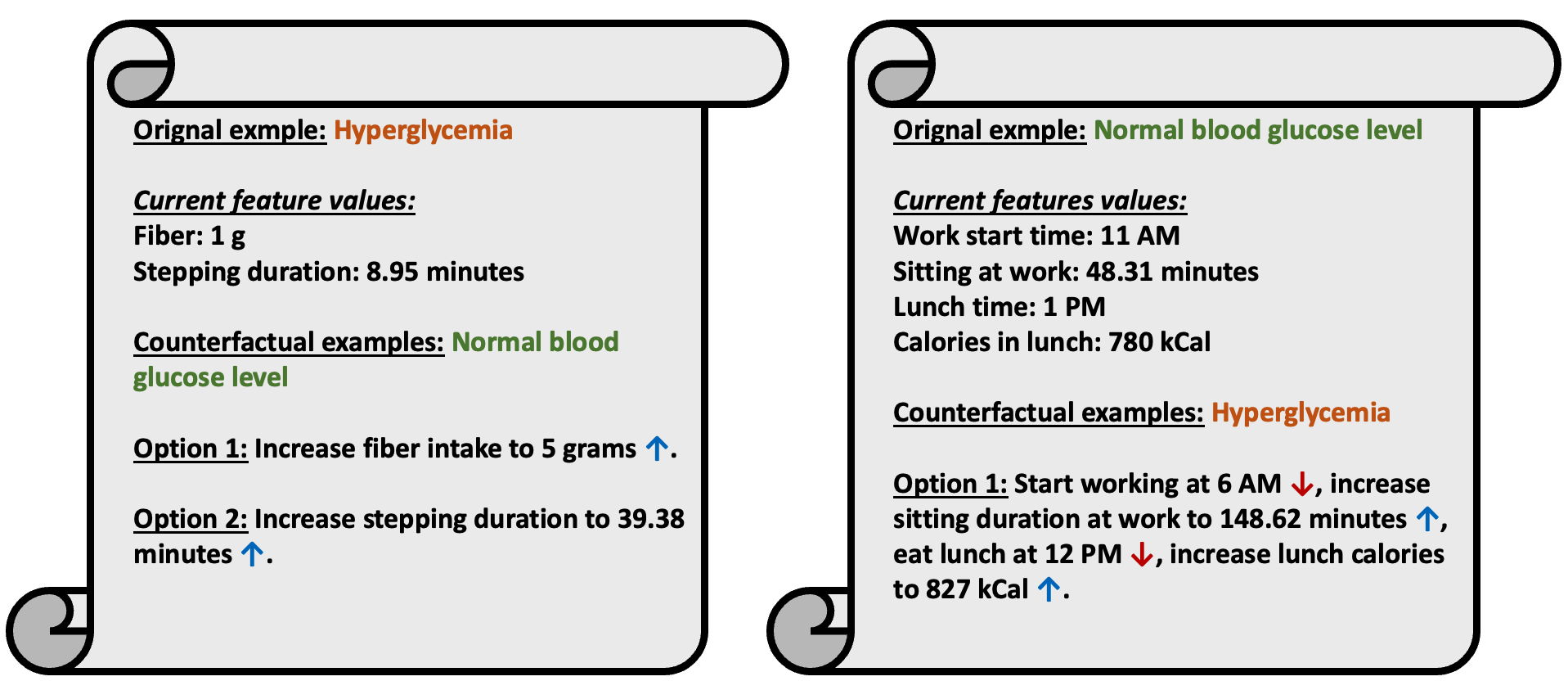}
    \caption{Counterfactual explanations for a normal and a postprandial hyperglycemic event. Features not presented in the figures had the same value in the original and all its counterfactual examples.}
\label{fig:cf_demo}
\end{figure}

\subsection{Results of Hyperglycemia Detection and Diverse Counterfactual Explanations}
\label{sec:hyperglycemia_results}
\begin{table}[htbp]
\centering
\caption{GlucoLens system's classification results with different pure and hybrid backbones for hyperglycemia detection on the 87\% training and 13\% test split. All metrics are averages over 100 repetitions with different random seeds.}
\begin{tabular}{lcccc}
\toprule
\textit{Classifier} & \textit{Accuracy}   & \textit{Precision}  & \textit{Recall}   & \textit{F1}    \\
\midrule
RF                  & 0.698          & 0.737          & 0.699          & 0.685          \\
XGB                 & 0.685          & 0.720          & 0.692          & 0.682          \\
MLP                 & 0.620          & 0.626          & 0.620          & 0.589          \\
RF+XGB              & 0.695          & 0.730          & 0.695          & 0.683          \\
RF+MLP              & 0.668          & 0.700          & 0.668          & 0.650          \\
XGB+MLP             & 0.687          & 0.712          & 0.687          & 0.672          \\
\textbf{RF+XGB+MLP} & \textbf{0.712} & \textbf{0.740} & \textbf{0.712} & \textbf{0.702} \\
\bottomrule
\end{tabular}
\label{tbl:different_classifiers}
\end{table}

\begin{table}[htbp]
\centering
\caption{Results of the GlucoLens hyperglycemia detection system with RF+XGB+MLP based hybrid backbone as we increase the training data size. All metrics are averages over 100 repetitions with different random seeds. An improvement of the performance metrics can be observed except for the last row, when only 1\% of the dataset is withheld for testing. In that case, the evaluation in any trial is vulnerable to producing 0 values for precision, recall, or F1 score, as there is only 1 example from each class in the test set. Eventually, it affects the overall average of those metrics.}
\begin{tabular}{lcccc}
\toprule
\textit{Size of   training set}  & \textit{Accuracy}   & \textit{Precision}  & \textit{Recall}   & \textit{F1}    \\
\midrule
70\% training, 30\% test         & 0.674          & 0.706          & 0.674          & 0.660          \\
80\% training, 20\% test         & 0.660          & 0.729          & 0.702          & 0.690          \\
87\% training, 13\% test     & 0.712          & 0.740          & 0.712          & 0.702          \\
90\% training, 10\% test         & 0.717          & 0.744          & 0.717          & 0.705          \\
\textbf{95\% training, 5\% test} & \textbf{0.733} & \textbf{0.751} & \textbf{0.733} & \textbf{0.716} \\
99\% training, 1\% test          & 0.730          & 0.625          & 0.730          & 0.660          \\
\bottomrule
\end{tabular}
\label{tbl:different_training_size}
\end{table}

GlucoLens achieves an accuracy of 73.3\% and an F1 score of 0.716 in hyperglycemia detection with a small dataset of 159 data points, and we prove with experiments that the performance improves as we increase the size of the dataset, as seen in \tblref{different_training_size}. RF, XGB, MLP, and their combinations were first trained and evaluated for a split of approximately 87\% training and 13\% test set. This split was created by randomly separating 10 positive class (hyperglycemia) examples and 10 negative class (normal) examples to create the test set. The rest of the examples were used to create the training set. Thus, the test set was already balanced, and the training set was balanced with ADASYN \cite{he2008adasyn}. The results for different classifiers can be seen in \tblref{different_classifiers}. We notice that the hybrid classifier that uses soft voting among an ensemble of RF, XGB, and MLP outperform the other combinations of classifiers. Note that the performance improvement in ensemble classifiers over individual classifiers are not due to the increased number of parameters because each model of the ensemble classifiers is trained independently (on the same training dataset), and only during inference on the test dataset, they work together to make more reliable predictions. We used the `All' feature set for the hyperglycemia detection experiments.

In the next part of the experiments, we show that as we increase the size of the training set, the performance keeps increasing. We varied the size of the training set from 70\% of the total dataset to up to 99\% of the dataset, and we noticed a strictly increasing pattern in all four metrics until the training set is 95\% of the dataset, as seen in \tblref{different_training_size}. This supports the fact that with a bigger dataset, the performance of our method can be much higher than what we are getting with this small dataset. For the last case, when 99\% of the dataset is used for training, the test set consists of only 1 example from each class. This case, makes the evaluation of the test set highly vulnerable for the precision and recall metrics, as unless the classifier predicts the whole test set perfectly in any trial, one of these two metrics would be zero for that trial, which would affect the F1 score as well. However, we believe that in a significantly larger dataset, even 1\% of the dataset could be used for testing. Overall, the best performance was found with RF+XGB+MLP with 95\%, 5\% train-test split with an accuracy of 73.3\% and a macro average F1 score of 0.716 with a precision of 0.751 and a recall of 0.733.

The visualization of counterfactual explanations in \figref{cf_demo} highlights insights for avoiding hyperglycemia by examining the influence of various features on postprandial blood glucose outcomes. The first example of \figref{cf_demo} represents a hyperglycemic case, characterized by a fiber intake of 1 gram and a stepping duration of 8.95 minutes. Counterfactual scenarios suggest pathways to achieve normal blood glucose levels. Option 1 illustrates that increasing fiber intake to 5 grams can lead to a normal outcome. Option 2 shows that substantially increasing stepping duration to 39.38 minutes also results in a normal blood glucose level. These scenarios emphasize the impact of dietary and physical activity modifications in managing glucose levels.

In contrast, a second example in \figref{cf_demo} represents a normal blood glucose level with current feature values of a work start time of 11 PM, sitting duration of 48.31 minutes, lunch time at 1 PM, and a lunch calorie intake of 780 kCal. Counterfactual scenarios reveal potential transitions to hyperglycemia. The counterfactual example generated against that example demonstrates how starting work earlier at 6 AM, increasing sitting duration to 148.62 minutes, shifting lunch to 12 PM, and increasing lunch calories to 827 kCal can contribute to hyperglycemia. These findings highlight the importance of balancing work habits, meal timing, and physical activity for glucose management, underscoring the value of counterfactual explanations for personalized recommendations.

\section{Discussion}
\subsection{Summary of the Results}
We have presented an important problem of estimating the postprandial area under the curve (AUC), the maximum postprandial blood glucose level (MaxBGL), and detecting hyperglycemia. Based on our experiments, we found that random forest (RF) models outperform multilayer perceptrons (MLP), ridge regression, XGBoost, and TabNet models in AUC prediction as shown in \tblref{model_comparison}. We also chose our features from five different feature sets as illustrated in \tblref{feature_sets}. Although our feature sets use 31 features as input to the model when choosing the set of `All' features, all these features are derived from only four sources of data, which makes it easier to have the required data easily available when needed for the model.

Our LLM-based experiments suggested that Claude Opus 4 outperforms GPT 3.5 Turbo, GPT 4, Gemini 2.0 Flash 001, DeepSeek V3, and Grok 3 in AUC predictions. We also found that hybrid models outperform LLM-only models, whereas augmented training data with Gaussian noise further improves the performance of the hybrid models. However, fully trainable base regressor backbones of our GlucoLens system outperform both LLM-based and hybrid versions of the predictors for AUC prediction.

Finally, in hyperglycemia prediction, we find that GlucoLens with ensemble classifiers with RF, XGB, and MLP outperform the system with individual classifiers, achieving an accuracy of 73.3\% and an F1 score of 0.716, as shown in \tblref{different_classifiers} and \tblref{different_training_size}. Additionally, we empirically prove that the GlucoLens hyperglycemia prediction system monotonically improves its test set performance as we increase the size of the training data.

\subsection{Different Interpretation of the Results}
Our best-found NRMSE of 0.123 implies that the predicted AUC value was on average within a 12.3\% error margin from the actual values of AUC. To interpret the performance in another way, we also explored the percentage of test cases falling within an error tolerance. From the actual AUC values and the predicted AUC values, the ratio of test cases within 5\%, 10\%, 15\%, and 20\% errors were calculated. Among the predictions made by our system, we have verified that 33\% of the cases had an error of less than 5\%, 61\% of the cases had an error of less than 10\%, 81\% of the cases had an error of less than 15\%, and 93\% cases had an error of less than 20\%. Thus only 19\% of the cases had an error of more than 15\%, and only 7\% of the cases had an error over 20\%. We believe that with more training data, our system will be able to predict most of the cases within a very small margin of error. The percentages of test cases within 5\%, 10\%, 15\%, and 20\% error margins for the XGBoost model were 30\%, 55\%, 78\%, and 92\%, respectively, all lower than our random forest model. Therefore, our random forest model outperforms XGBoost in this metric as well.

\subsection{Alternative Feature Search Methods}
We also explored another avenue of research to find possible feature sets other than the five options presented in \tblref{feature_sets}. One way could be to perform an exhaustive search within the features. The problem with that approach would be the computational cost. As we have 31 features, we would have $2^{31}$ possibilities which would be infeasible to explore. However, we have done an additional experiment to find effective feature subsets. Among all the models, our random forest (RF) was the best model so far with an NRMSE of 0.123. For finding alternative subsets, we limited the number of features by restricting the number of leaf nodes in RF models to 24, 48, and 96. The 48-leaf-node limit outperformed the others, achieving an NRMSE of 0.121, surpassing our previous best result of 0.123. This approach offers a more rational feature selection method, as limiting leaf nodes helps identify features with higher information gains, reduces overfitting, and improves generalization.

\section{Conclusion}
In this study, we have developed an interpretable ML system for estimating postprandial blood glucose parameters by analyzing different diet and physical activity features. We believe that our tools will be helpful to people diagnosed with or at risk of diabetes to better regulate their condition and avoid unwanted outcomes. We developed a multi-sensor-based system, GlucoLens, that is based on advanced data processing for creating multimodal feature sets and machine learning regressors and classifiers, and large language models for hyperglycemia prediction. Based on the experiments with different variations of models and feature sets, our GlucoLens AUC prediction system in its best configuration outperforms the comparison models by a margin of 10\% to 27\%. Moreover, it achieves a 73.3\% accuracy and an F1 score of 0.716 on hyperglycemia prediction. Finally, it recommends different interventions to avoid hyperglycemia based using diverse counterfactual explanations. Based on the explanations, we showed that a regulated diet and higher physical activity can avoid hyperglycemia.


\reftitle{References}




\bibliography{root}

\end{document}